\documentclass{article}

\usepackage{arxiv}

\usepackage[utf8]{inputenc} 
\usepackage[T1]{fontenc}    
\usepackage{hyperref}       
\usepackage{url}            
\usepackage{booktabs}       
\usepackage{amsfonts}       
\usepackage{amsmath}        
\usepackage{nicefrac}       
\usepackage{microtype}      
\usepackage{lipsum}		
\usepackage{graphicx}
\usepackage{natbib}
\usepackage{doi}

\usepackage{subcaption} 

\title{Reduced-Order Model-Guided Reinforcement Learning for Demonstration-Free Humanoid Locomotion}

\author{ 
    Shuai Liu\\
	Laurentian University\\
	\texttt{sliu16@laurentian.ca} \\
	\And
    Meng Cheng Lau \\
	Laurentian University\\
	\texttt{mclau@laurentian.ca} \\
}

\renewcommand{\shorttitle}{}

\hypersetup{
pdftitle={Reduced-Order Model-Guided Reinforcement Learning for Demonstration-Free Humanoid Locomotion},
pdfsubject={cs.RO, cs.LG},
pdfauthor={Shuai Liu, Meng Cheng Lau},
pdfkeywords={},
}

\begin{document}
\maketitle

\begin{abstract}
We introduce Reduced-Order Model-Guided Reinforcement Learning (ROM-GRL), a two-stage reinforcement learning framework for humanoid walking that requires no motion capture data or elaborate reward shaping. In the first stage, a compact 4-DOF (four-degree-of-freedom) reduced-order model (ROM) is trained via Proximal Policy Optimization. This generates energy-efficient gait templates. In the second stage, those dynamically consistent trajectories guide a full-body policy trained with Soft Actor–Critic augmented by an adversarial discriminator, ensuring the student’s five-dimensional gait feature distribution matches the ROM’s demonstrations. Experiments at 1 m/s and 4 m/s show that ROM-GRL produces stable, symmetric gaits with substantially lower tracking error than a pure-reward baseline. By distilling lightweight ROM guidance into high-dimensional policies, ROM-GRL bridges the gap between reward-only and imitation-based locomotion methods, enabling versatile, naturalistic humanoid behaviors without any human demonstrations.
\end{abstract}

\begin{center}
\href{https://youtu.be/aRsLgLePXdE}{\textbf{Supplemental video:} https://youtu.be/aRsLgLePXdE}
\end{center}


\section{Introduction}
Achieving natural humanoid locomotion is a longstanding goal in both robotics and computer animation. From bipedal robots that can walk and run with human-like grace, to virtual characters that move realistically in games, the ability to synthesize lifelike walking gaits remains a persistent challenge. Traditional model-based control has produced impressive feats, but often requires painstaking design and does not always capture the natural motion nuances of human walking \cite{kuindersma_optimization-based_2016}. In recent years, reinforcement learning (RL) has emerged as a promising data-driven paradigm for developing locomotion controllers \cite{radosavovic_real-world_2024}. RL allows simulated humanoids to learn complex gait behaviors through trial-and-error, offering the potential to discover agile and robust walking strategies that would be difficult to manually design.

Reinforcement learning (RL) methods for natural locomotion generally fall into two main paradigms aimed at producing lifelike gait behaviors.

A purely objective‑driven RL policy can discover stable walking gaits by optimizing energy and stability rewards. These reference-free methods train locomotion policies from scratch by optimizing carefully crafted reward functions, without any motion capture examples. The reward terms are designed to encourage physically plausible and human-like traits, such as forward speed with energy efficiency, maintaining center-of-mass stability and upright posture, periodic foot contact patterns, and symmetric gait cycles between left and right legs. By rewarding such objectives, controllers can spontaneously develop stable walking gaits that emerge naturally from learning. Notably, researchers have demonstrated that symmetric bipedal walking can arise solely from reward design and curriculum training, without any reference motions \cite{yu_learning_2018}. Indeed, recent work showed that an RL policy trained with biomechanically inspired rewards could produce natural walking behaviors purely through self-exploration \cite{peng_gait-conditioned_2025}. The appeal of this approach is that it does not require any pre-recorded motions – the agent invents its own walking cycle. However, designing the reward function is notoriously difficult: the policy’s behavior is highly sensitive to the choice and weighting of reward terms, often requiring elaborate hand-tuning and expertise. Even with careful tuning, purely objective-driven policies may develop subtle artifacts or unnatural quirks since there is no direct template of “human” motion to imitate. In summary, while motion-free RL can yield impressively natural gaits under the right conditions, it faces challenges in reward engineering and consistency of motion style.

Alternatively, imitation‑driven RL uses human mocap examples to ensure motion realism but at the cost of dataset dependency. To directly ensure natural movement quality, a dominant approach is to imitate example locomotion trajectories from motion capture data. In this paradigm, the RL agent is guided by reference motions of humans (or animals) and receives rewards for matching the reference pose and velocity at each time step, or uses adversarial critics to judge realism against a motion dataset \cite{10.1145/3197517.3201311,10.1145/3450626.3459670}. DeepMimic pioneered this line of work by showing that standard RL algorithms can learn robust control policies capable of imitating a broad range of example motion clips from a motion capture library \cite{10.1145/3197517.3201311}. By combining motion imitation objectives with task goals, DeepMimic enabled physically simulated characters to reproduce dynamic skills (flips, spins, walks) with high fidelity to the mocap examples. Following this, numerous studies have pushed the state of the art in imitation-based locomotion. For instance, ASE (Adversarial Skill Embeddings) uses large unstructured motion datasets to train latent skill models via adversarial imitation, yielding a repertoire of reusable behaviors that look remarkably life-like \cite{10.1145/3528223.3530110}. Another example is CALM (Conditional Adversarial Latent Models), which learns a rich latent representation of human movement through imitation learning, capturing the complexity and diversity of human motion while allowing direct user control over the character’s style and direction \cite{Tessler_2023}. Most recently, the concept of Behavioral Foundation Models have emerged – here a policy is pre-trained on massive collections of motion data (in an unsupervised manner) to serve as a generalist locomotion model. These foundation models, once trained on unlabeled motion trajectories, can be prompted to perform new tasks in a zero-shot fashion, while retaining human-like gait qualities \cite{tirinzoni2025zeroshot}. Motion imitation approaches thus achieve state-of-the-art realism in simulated walking; policies closely mimic human kinematics and can produce motions nearly indistinguishable from motion capture. The downside, however, is their heavy reliance on curated motion data – one must have access to large datasets of reference gaits, and the learned skills are inevitably tied to the distribution of motions in those datasets.

Despite these advances, achieving natural, human‑like locomotion without any motion capture data remains an open challenge. Purely objective‑driven methods often fail to reproduce the fluidity and subtle timing of human gait, while imitation‑based approaches are fundamentally constrained by the availability and diversity of mocap archives.

To bridge this gap, we introduce a reduced‑order model–guided reinforcement learning(ROM-GRL) framework. It leverages a simplified locomotion model as a stand‑in for motion capture, providing high‑level gait guidance to the RL agent without any human demonstrations. Our framework unfolds in two stages. In the first stage, we train a lightweight teacher model to generate efficient, dynamically consistent gait templates that capture the essence of natural walking. In the second stage, we distill these templates into a full-body controller by rewarding adherence to the teacher’s motion distribution, ensuring smooth, human-like locomotion. By separating high-level gait planning from detailed control, we achieve a single walking policy that is both robust and naturally fluid, all without relying on motion capture data or intricate reward design.

In summary, our approach marries the insights from model-based gait generation with the flexibility of reinforcement learning. By using a ROM to guide RL instead of direct motion capture, we maintain a purely physics-driven training regime while still inducing realistic movement patterns. The proposed framework demonstrates that natural humanoid locomotion can emerge without demonstrations, closing the gap between reward engineering and motion imitation. We validate that our ROM-guided RL method produces walking controllers that are stable across different speeds and exhibit natural gait symmetry and fluidity comparable to reward-based policies, suggesting a novel solution to produce life-like humanoid locomotion in the absence of motion data.

\section*{Declarations}
Portions of this manuscript were drafted with the assistance of an AI language model. All scientific content, interpretations, and final edits were performed by the human authors.

\section{Related Work}
\label{sec:headings}

\subsection{Humanoid Locomotion}
Humanoid locomotion research draws inspiration from the natural gaits of humans and animals, which are both efficient and versatile. Early bipedal robots achieved passive dynamic walking, demonstrating that a simple unpowered mechanism can walk stably on a slight slope by exploiting gravity and inertia \cite{gu2025humanoidlocomotionmanipulationcurrent}. These passive walkers highlighted how mechanical dynamics alone can produce human-like gait cycles with minimal energy input. Based on this idea, engineers have created controllers that focus on energy efficiency by adding compliance and recovering energy so that humanoid robots can approach the metabolic economy of human walking. Over the decades, bipedal locomotion progressed from passive or quasi-static gait strategies to fully dynamic walking as control techniques improved \cite{gu2025humanoidlocomotionmanipulationcurrent}. On flat terrain, periodic walking motions are now well understood and can be reliably generated with model-based methods. Beyond steady-state walking, researchers have expanded toward more agile motion: modern humanoids are expected not only to walk, but also to run, jump, and handle disturbances with athletic skill. Classical model-based approaches (e.g. zero-moment point trajectory planners) enabled some early agility, but recent learning-based controllers have dramatically advanced the state of the art. For example, reinforcement learning (RL) policies have learned to make a humanoid traverse stairs and uneven ground without vision \cite{wu2025learnteachsampleefficientprivileged}, and to perform dynamic maneuvers like leaping forward or recovering from large pushes \cite{gu2025humanoidlocomotionmanipulationcurrent}. Other work has even applied deep RL to teach humanoid robots high-speed kicks and complex behaviors inspired by sports, such as agile soccer motions including rapid fall recovery, significantly faster walking, and agile turning \cite{haarnoja_learning_2024}.

\subsection{Reinforcement Learning}

Deep reinforcement learning has become a powerful tool for developing control policies in high-dimensional systems like humanoids. To tame the complexity of locomotion, researchers often adopt a hierarchical reinforcement learning approach that breaks the task into layered sub-problems. In a typical hierarchy, a high-level (HL) policy plans abstract actions, such as selecting footsteps, velocities, or gait patterns – and a low-level (LL) controller then executes the details of joint coordination to realize those plans. Such frameworks can combine learning-based policies with traditional controllers: for instance, a learned high-level policy can command a model-based balancing controller \cite{bao2025deepreinforcementlearningbipedal}. This modular design leverages abstraction (the HL deals with coarse objectives, the LL with fast dynamics), which can improve sample efficiency and robustness by incorporating prior knowledge at the appropriate level.

Unconstrained RL optimizations may find physically correct but unnatural-looking gaits (for instance, overly stiff or asymmetric strides) that satisfy the reward but would be suboptimal or unsafe for a real robot. Traditional humanoid control relied on expert-designed reference motions and gait libraries to guarantee natural behavior, which required significant human insight and tuning. To reduce this manual effort, newer approaches incorporate reference motion data or demonstrations into the learning process \cite{bao2025deepreinforcementlearningbipedal}. By guiding the policy toward known-good motion patterns, these methods bias learning towards natural gaits. For example, recent work has combined RL by tracking human walking motion to produce controllers that exhibit human-like locomotion \cite{bao2025deepreinforcementlearningbipedal}. The RL policy in that framework is rewarded for following the reference trajectory, resulting in smoother, more life-like movements. By penalizing large changes in motion or deviations from expected gait distributions, such constraints help maintain stability and realism.

\subsection{Reduced-Order Model}
One promising direction to incorporate domain knowledge into locomotion learning is through reduced-order models (ROMs) of the dynamics. ROMs are simplified representations of legged locomotion, capturing the essential physics of walking or running with only a few abstract variables. Classic examples include the Linear Inverted Pendulum Model (LIPM) \cite{1241826} for walking, which simplifies a biped to a point mass moving over a support leg, and the Spring-Loaded Inverted Pendulum (SLIP) model for running \cite{blickhan1989spring}, which treats the stance leg like a spring.
These models have a long history in humanoid control: they are computationally lightweight and often reflect the “natural” dynamics of gait, making them ideal for high-level planning. Many model-based controllers use a ROM (such as a centroidal momentum model or inverted pendulum) to plan center-of-mass trajectories and foot placements, which are then realized by a whole-body controller \cite{gu2025humanoidlocomotionmanipulationcurrent}. The success of ROMs in planning suggests that they capture fundamental gait principles – a fact that our approach leverages in the context of learning.

In recent years, researchers have begun using simplified teacher dynamics from ROMs to guide reinforcement learning agents. The idea is to marry the intuition of a reduced-order model with the flexibility of RL: the ROM provides high-level guidance or demonstration trajectories, and RL handles the low-level control needed for the full humanoid. Prior work by Green et al. exemplifies this strategy, using a spring-mass running model as the high-level planner for a bipedal robot \cite{9380929}. In their framework, the ROM generates a desired motion (e.g. target foot positions and timings based on the spring-mass dynamics), which serves as a command and training signal for the low-level policy. Rather than tracking the ROM trajectory exactly, the learned policy is trained to balance immediate tracking with long-term stability, compensating for differences between the idealized model and the real robot \cite{9380929}. This approach effectively uses the ROM as an expert demonstrator or teacher: it encapsulates expert knowledge of locomotion (like how a stable gait should look in abstract terms), and the RL agent learns to reproduce that behavior on the complex, high-dimensional humanoid. More generally, using a ROM in this way constrains the policy search to a smaller meaningful subspace of behaviors. Instead of exploring arbitrary joint motions, the learning process is biased toward physically plausible motions that the ROM would produce. This significantly improves sample efficiency and can yield more robust gaits, since the agent is effectively taught the “tricks” of balancing and momentum management from a simple model. Building on these insights, our method integrates a reduced-order teacher model into the training loop, aiming to combine the ROM’s inherent agility and efficiency with the adaptive power of reinforcement learning. The following sections will detail how this integration is realized, and how it addresses the gaps identified in prior work.

In summary, ROMs have proven invaluable for guiding complex locomotion learning, from spring-mass models for walking \cite{9380929} to point-mass models for brachiation \cite{10.1145/3528233.3530728}. Our work builds directly on this insight. We distinguish ourselves by how the ROM guidance is used: instead of hard trajectory tracking or post-hoc initializations, we embed the ROM’s behavior into the RL training objective via distribution constraints. This ensures the student policy not only follows the teacher’s footprints but statistically behaves like the reduced model in terms of gait dynamics. Consequently, our humanoid policy inherits the ROM’s stability and efficiency properties, yet is free to adjust its joint actuation to handle high-dimensional nuances. This ROM-to-policy distillation with distributional fidelity is a novel contribution beyond prior approaches based on simplified models. By doing so, we achieve natural and agile locomotion through a principled transfer of simplified dynamics: the ROM provides the blueprint of efficient movement, and the RL student brings in the full-body realism, with both aligned through our distribution-constrained teacher–student training.




\section{Methodology}

\begin{figure}[htbp]
  \centering
  \includegraphics[width=\textwidth]{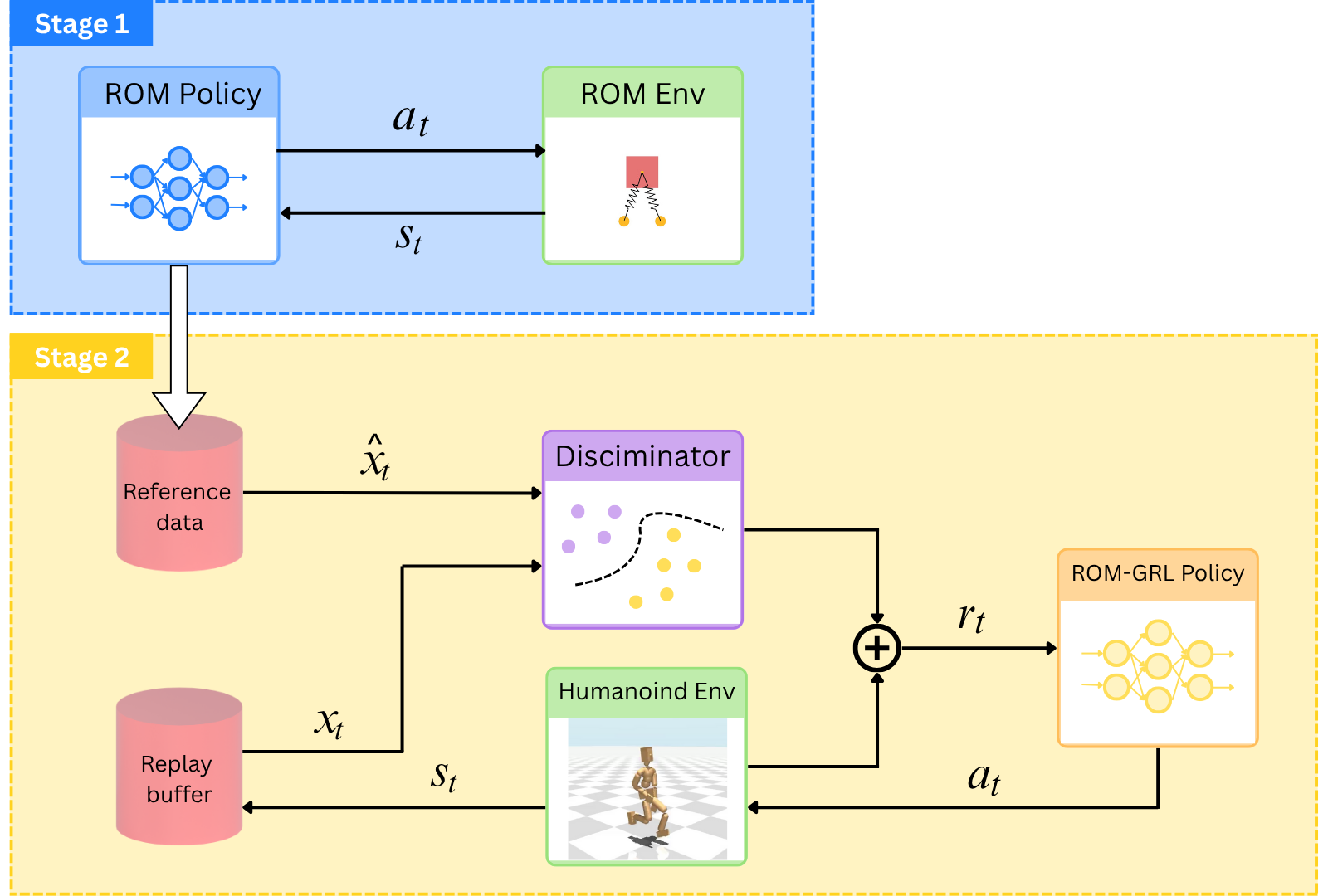}
  \caption{Overview of the ROM-GRL framework. In Stage~1, a 4-DOF ROM policy is trained in Box2D: the policy interacts with the environment and the resulting key gait signals are recorded as reference data \(\hat{x}_t\). In Stage~2, both the reference signals \(\hat{x}_t\) and the student’s live feature vector \(x_t\) (extracted from the Humanoid environment) are fed into an adversarial discriminator, whose output defines an imitation bonus. This bonus is added to the environment reward \(r_t\) to train the ROM-GRL policy via Soft Actor–Critic, while all transitions are stored in a replay buffer. The resulting policy produces robust, natural locomotion that closely matches the ROM’s gait feature distribution.}
  \label{fig:overview_pipeline}
\end{figure}

We introduce Reduced-Order Model–Guided Reinforcement Learning (ROM-GRL), a two-stage training framework that leverages the simplicity of a compact dynamic model to generate dynamically consistent gait representations and then distills those representations into a high-dimensional walking controller via a soft distributional imitation constraint, encouraging statistical similarity to teacher trajectories. By first producing energetically efficient reference motions and then enforcing adherence to the 5-D gait feature distribution during full-body policy learning, ROM-GRL yields a single, robust locomotion policy that exhibits natural gait dynamics without motion capture data or elaborate reward engineering. Fig.~\ref{fig:overview_pipeline} outlines our framework.

\subsection{Overview of the Two-Stage Framework}

Our framework consists of two stages.

(1) \textbf{Trajectory Generation via ROM:} We first generate reference locomotion trajectories using a planar, 4-DOF reduced-order dynamic model of bipedal walking. This model encodes essential gait dynamics, including center-of-mass oscillation, pendulum-like leg swing, and periodic footfall timing, and can produce energetically efficient, balanced walking motions through simple forward-velocity optimization. We train a teacher policy on this ROM, which converges rapidly due to the low dimensionality, then roll it out for 2,000 time steps to record the resulting trajectories of the body center and each foot. These ROM-produced signals serve as dynamically consistent representations of natural walking, without relying on any human data.

(2) \textbf{Teacher–Student RL with Distribution Constraint:} Next, we distill the ROM-guided behaviors into a full-body controller. In the teacher phase, the ROM-trained policy provides high-quality gait executions for a specified target speed. In the student phase, we train a full-body policy to maximize the standard primary reinforcement objective (e.g., forward velocity) while also matching the teacher’s 5-D gait feature distribution. We implement this via an adversarial discriminator that learns to distinguish between ROM-generated and student-generated trajectory snippets, and use its output to compute a soft imitation bonus. By encouraging similarity to the ROM’s distribution, the student policy preserves natural gait characteristics and avoids degeneration into unstable or unnatural patterns while still optimizing for task performance.

Both phases employ the same simple forward-velocity reward and require no additional reward shaping or motion capture data. The result is a single walking policy that is robust, efficient, and naturally fluid.

\begin{figure}
    \centering
    \includegraphics[width=0.5\linewidth]{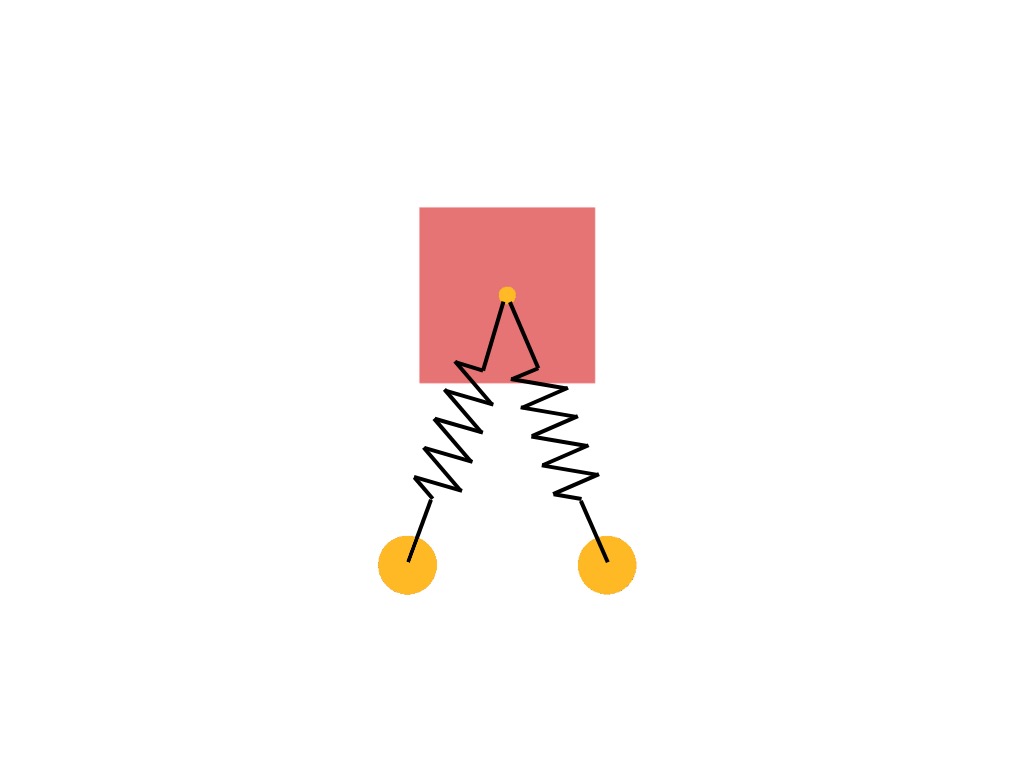}
    \caption{Schematic of the planar ROM used to generate reference walking trajectories. The ROM consists of a central mass representing the body’s center of mass, connected to two point feet via massless linear springs and actuated revolute hip joints.}
    \label{fig:ROM}
\end{figure}

\subsection{Stage 1: ROM Trajectory Generation and Teacher Policy Training}
\label{sec:stage1}

We implement a planar ROM, illustrated in Fig.~\ref{fig:ROM}, to generate compact and physically plausible walking reference trajectories. The ROM abstracts the target humanoid into a 4-DOF (four-degree-of-freedom) mechanism: a central square of mass \(m\) representing the system’s center of mass, two massless linear springs connecting it to point feet via revolute hip joints, and continuous torque actuation at each hip joint. This simple structure captures pendulum-like leg dynamics, center-of-mass oscillation, and periodic footfall timing, while remaining low-dimensional enough to train rapidly. We expose the ROM as a Gymnasium \cite{towers2024gymnasium} environment whose 20-dimensional observation vector includes the body’s planar position and velocity, each spring’s current length and extension velocity, both hip joint angles and angular velocities, and the target forward speed; its two-dimensional action applies continuous torques at the left and right hips; and its reward is defined as
\[
  r_t = \exp\bigl(-\alpha\,\lvert v_{\mathrm{COM}} - v_{\mathrm{target}}\rvert\bigr),
\]
where \(v_{\mathrm{COM}}\) is the horizontal velocity of the central mass, \(v_{\mathrm{target}}\) is the commanded speed, and \(\alpha\) scales the sensitivity, yielding a bounded reward in \((0,1]\). We train a teacher policy on this ROM using Proximal Policy Optimization (PPO) \cite{schulman2017proximalpolicyoptimizationalgorithms} until convergence. Upon convergence, we roll out the trained policy for \(T\) time steps and record 5-D trajectory snippet at each step:
\[
  y_{\mathrm{COM}}(t),\quad
  \bigl(x_{L}(t),\,y_{L}(t)\bigr),\quad
  \bigl(x_{R}(t),\,y_{R}(t)\bigr),
  \quad t=1,\ldots,T.
\]
where \(y_{\mathrm{COM}}(t)\) is the vertical position of the body’s center of mass at time \(t\),  while \( x_{L}(t) \), \( y_{L}(t) \), \( x_{R}(t) \), and \( y_{R}(t) \) represent the horizontal and vertical positions of the left and right feet, measured relative to the center of mass. We select these signals because the vertical motion of the center of mass \(y_{\mathrm{COM}}\) captures the global balance dynamics, while the planar coordinates of the left and right feet \((x_{L},y_{L})\) and \((x_{R},y_{R})\) encapsulate foot placement, stride length, and periodic timing, together providing a minimal yet expressive representation for natural gait.

\subsection{Stage 2: Student Policy Training}
\label{sec:stage2}

When trained solely on the forward-velocity reward, policies often exhibit jerky or unstable gaits, and their gait becomes unpredictable and uncontrollable. To enforce smooth, human-like locomotion, we impose a soft distributional constraint derived from the ROM's reference trajectories. An adversarial discriminator \(D\) is trained to distinguish between “real” snippets sampled from the ROM data and “fake” snippets generated by the student policy. Each trajectory segment
\[
  \tau = [\,y_{\mathrm{COM}},\,x_{L},\,y_{L},\,x_{R},\,y_{R}\,]
\]
is judged by \(D\), which minimizes the binary cross‐entropy loss (as in standard GANs \cite{goodfellow2014generative}):
\[
  \mathcal{L}_D = -\mathbb{E}_{\tau\sim\mathrm{ROM}}\bigl[\log D(\tau)\bigr]
                -\mathbb{E}_{\tau\sim\pi}\bigl[\log(1 - D(\tau))\bigr].
\]
This loss consists of two parts: the first term encourages \(D\) to assign high probability \(D(\tau)\) to “real” snippets drawn from the ROM (by minimizing \(-\log D(\tau)\)), while the second term encourages \(D\) to assign low probability (i.e.\ \(D(\tau)\) close to zero) to “fake” snippets generated by the student policy (by minimizing \(-\log(1 - D(\tau))\)). Together, these terms train \(D\) to distinguish ROM trajectories from student trajectories by maximizing the likelihood of correct labels under a binary classification framework.
During student rollouts, each segment \(\tau\) incurs an imitation bonus
\[
  r_{\mathrm{im}}(\tau) = -\log\bigl(1 - D(\tau)\bigr),
\]
which is large when the discriminator considers \(\tau\) likely under the ROM distribution. To optimize our high-dimensional student policy, we adopt Soft Actor–Critic (SAC) as the base reinforcement learning algorithm. SAC is an off-policy actor–critic method that augments the standard expected return objective with an entropy term, encouraging exploration and preventing premature convergence to suboptimal deterministic behaviors~\cite{haarnoja2018softactorcriticoffpolicymaximum}. Its sample efficiency and robustness to hyperparameter choices make it well‐suited for complex control tasks like full-body locomotion, where both stability and adaptability are critical. We train the student policy \(\pi\) with Soft Actor–Critic to maximize the combined objective
\[
  J(\pi) = \mathbb{E}_{(s,a)\sim\pi}\bigl[\eta\,r_{\mathrm{env}}(s,a)\bigr]
          + \mathbb{E}_{\tau\sim\pi}\bigl[(1-\eta)\,r_{\mathrm{im}}(\tau)\bigr],
\]
where \(r_{\mathrm{env}}\) is the standard velocity-matching reward and \(\eta=0.5\) balances task performance with imitation fidelity. This integrated approach ensures the student policy not only achieves the target speed but also adheres closely to the natural gait characteristics encoded by the ROM.

In summary, our methodology employs a two-stage framework: Stage 1 uses a 4-DOF ROM to produce five key trajectory signals as dynamically consistent gait templates, and Stage 2 distills those templates into a high-dimensional full-body policy via an adversarial distribution constraint within SAC. This design yields a robust, fluid walking controller without recourse to motion capture data or elaborate reward engineering.

\section{Experimental Results}

\subsection{Experimental Setup}

\subsubsection{Stage 1: ROM Trajectory Generation and Teacher Policy Training}

Our planar ROM is implemented in C++ using Box2D \cite{catto2010box2d} and exposed to Python via pybind11 \cite{pybind11}. We wrap this implementation in a Gymnasium environment to achieve high step‐throughput for training, leveraging the efficiency of the C++ backend. The environment’s 20-dimensional observation vector comprises body-center positions and velocities, spring lengths and velocities, hip angles and angular velocities, while its two-dimensional action space applies continuous torques in \([-1,1]^2\) to the left and right hip joints.

To generate both open-loop reference trajectories and a learned teacher policy, we train using the Sample Factory \cite{petrenko2020sf} implementation of Asynchronous PPO (APPO). The policy and value networks each consist of a two-layer multilayer perceptron (MLP), each with 512 hidden units and ReLU activations. We optimize with Adam (learning rate \(1\times10^{-4}\)), a discount factor \(\gamma=0.99\), and a clipping ratio of 0.2, collecting experience from eight parallel actors over one million environment steps. At each time step, the reward is simply the forward (positive \(x\)) component of the center-of-mass velocity. Under this setup, the teacher converges to a stable walking gait in approximately \(5\times10^5\) steps. We record all rollout trajectories to construct the reference dataset for Stage 2 student policy distillation.

\subsubsection{Stage 2: Policy Training}

In Stage 2, we build on the HumEnv benchmark~\cite{tirinzoni2024metamotivo}, modifying the underlying MuJoCo model to enforce strictly sagittal-plane locomotion by fixing both lateral translation and yaw rotation at the root. This restriction significantly reduces the system’s degrees of freedom and simulation complexity, enabling more efficient training under limited computational resources. The resulting observation space is 358-dimensional and includes the root height, as well as the local position, rotation (in 6D representation), linear velocity, and angular velocity of each of the 24 rigid body segments. All features are expressed in the agent's body frame. The action space is a 21-dimensional vector of hinge-joint torques, each normalized to the interval \([-1,1]\), one per each of the 21 actuated revolute joints in the model.

We implement our ROM-guided reinforcement learning algorithm using Soft Actor–Critic (SAC) from the CleanRL library \cite{huang2022cleanrl}, with an adversarial discriminator that encourages the agent consisting of the root height and the (x, z) positions and velocities of both ankles (total of five dimensions), to match those of the pre-trained ROM teacher. The discriminator is a multilayer perceptron with two hidden layers of 64 and 32 units, respectively, each followed by LeakyReLU activations , with dropout \(p=0.2\) applied after each layer. During training, we inject Gaussian noise (\(\sigma=0.05\)) into both expert and agent features and apply label smoothing (0.9 for expert labels, 0.1 for agent labels). We optimize the discriminator with Adam (learning rate \(1\times10^{-5}\)) and include a gradient penalty term (coefficient \(\lambda=10\)) computed on random interpolations of expert and agent samples to enforce a 1-Lipschitz constraint. The discriminator is updated once every five SAC actor–critic updates, beginning at global step 5{,}000, and training is terminated early if validation loss in a holdout 10\% subset of ROM trajectories fails to improve for ten consecutive updates.

\subsection{Results}

We evaluate our ROM‐GRL method against a pure‐reward RL baseline, a full‐body MuJoCo policy trained with Soft Actor–Critic using only the forward‐velocity environment reward, at two target speeds: 1 m/s and 4 m/s. The ROM teacher was trained for a target speed of 4 m/s, so quantitative comparisons at that speed yield a meaningful assessment of tracking fidelity. To demonstrate generality, we also train the full‐body policy at 1 m/s and include qualitative results at both speeds. However, due to the fundamentally different dynamics at lower speeds, the 1 m/s policy shows larger discrepancies in quantitative error metrics compared to the ROM teacher; accordingly, we report quantitative analysis only at 4 m/s, while presenting qualitative trajectory comparisons at both speeds.

\subsubsection{Qualitative Trajectory Comparison}

\begin{figure}[htbp]
  \centering
  \begin{subfigure}[b]{0.48\textwidth}
    \centering
    \includegraphics[width=\linewidth]{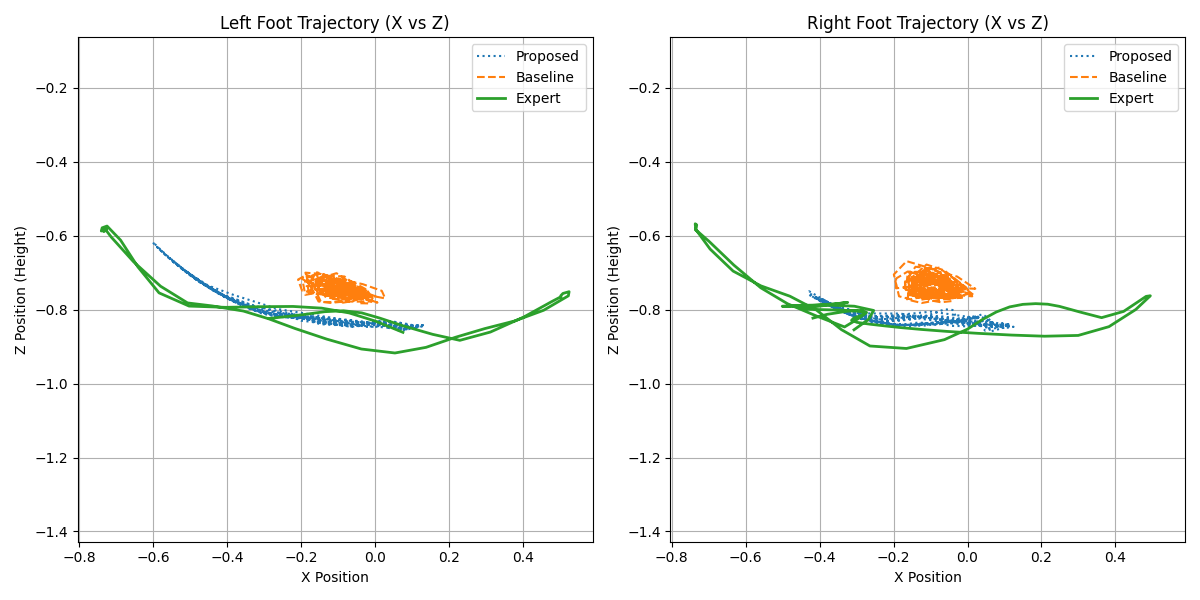}
    \caption{1 m/s}
    \label{fig:proposed-1mps}
  \end{subfigure}
  \hfill
  \begin{subfigure}[b]{0.48\textwidth}
    \centering
    \includegraphics[width=\linewidth]{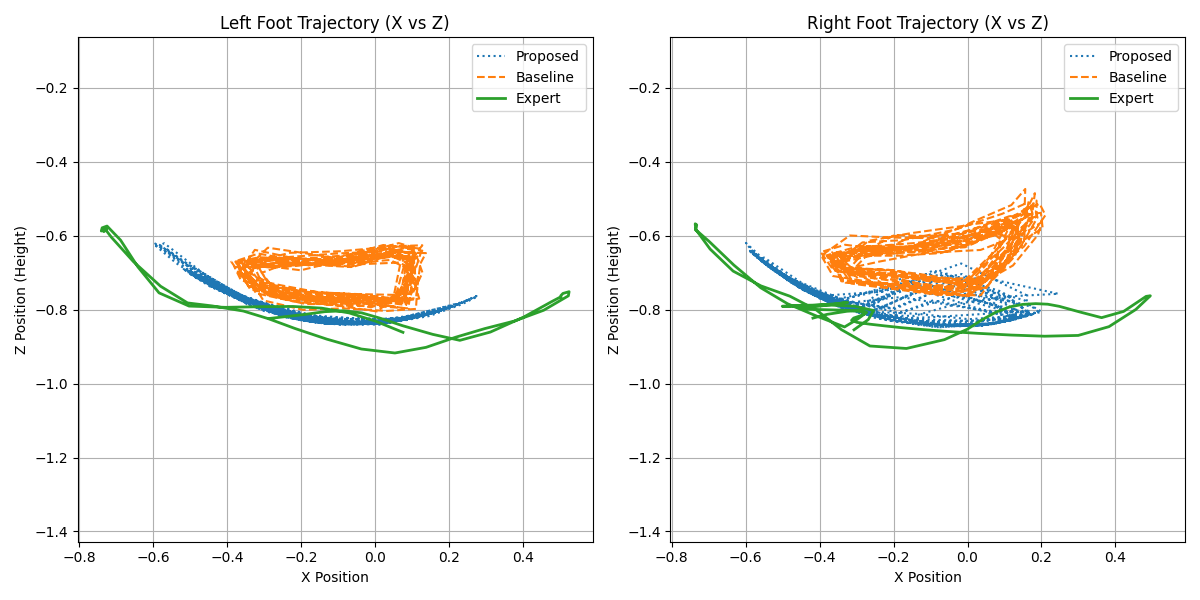}
    \caption{4 m/s}
    \label{fig:proposed-4mps}
  \end{subfigure}
  \caption{Comparison of pelvis and foot trajectories for (a) 1 m/s and (b) 4 m/s. Curves show the ROM teacher (green), ROM-GRL policy (blue), and pure-reward RL baseline (orange).}
  \label{fig:baseline-comparisons}
\end{figure}

Figure~\ref{fig:baseline-comparisons} visualizes pelvis and foot trajectories for the ROM-GRL policy (blue) and the pure-reward baseline (orange), both overlaid with the ROM teacher reference (green). (a) shows 1 m/s results, and (b) shows 4 m/s results. At both speeds, our policy tracks the teacher more closely during swing and stance phases, exhibiting improved foot clearance and gait symmetry. The baseline, in contrast, deviates substantially during swing, leading to lower foot height and asymmetric strides.

\subsubsection{Temporal Tracking Performance at 4 m/s}

\begin{figure}[htbp]
  \centering
  \includegraphics[width=0.75\linewidth]{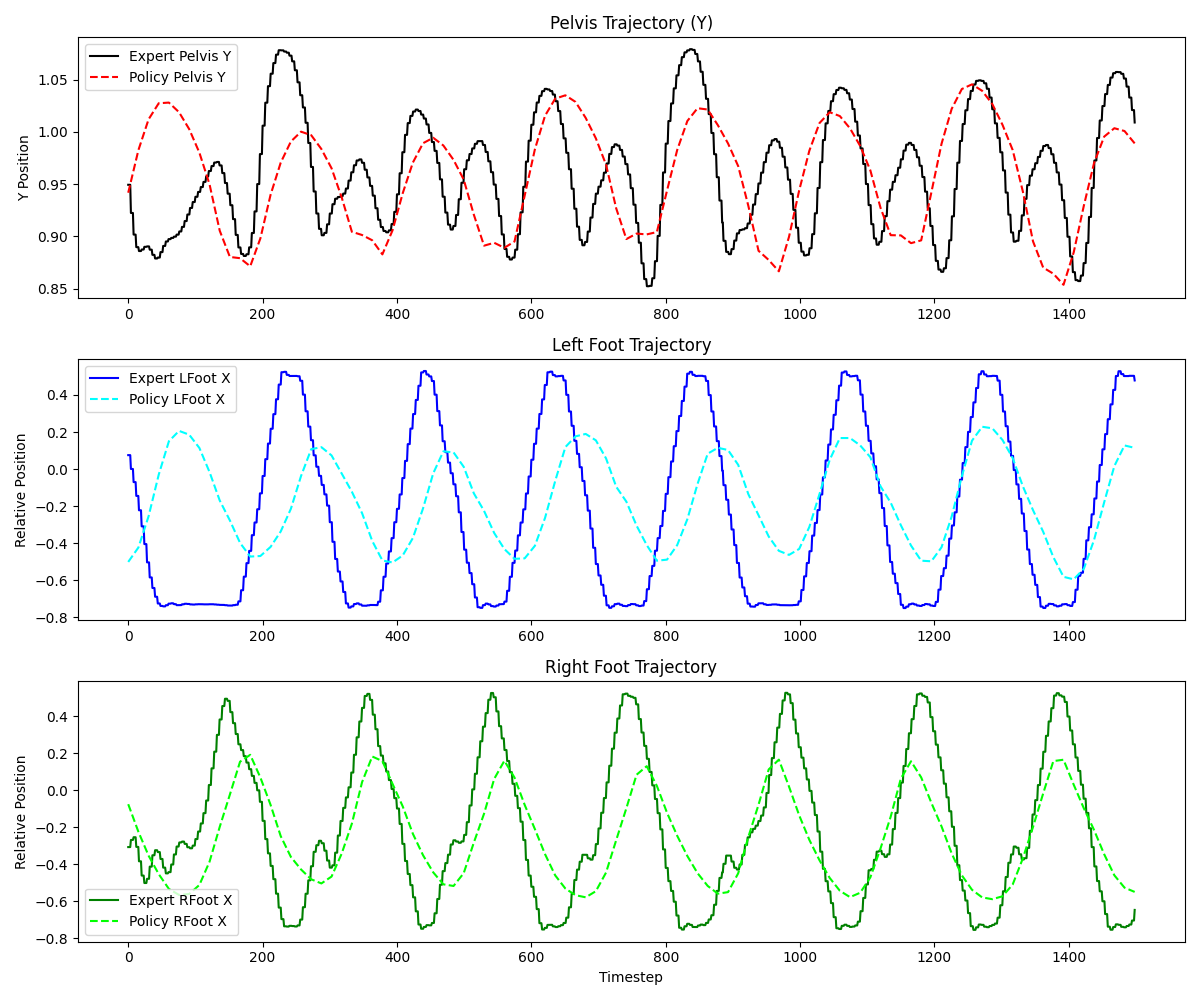}
  \caption{Time‐domain comparison at 4 m/s: pelvis Y (top), left foot X (middle), right foot X (bottom). ROM teacher (solid) vs.\ ROM‐GRL policy (dashed).}
  \label{fig:temporal-trajectories-4ms}
\end{figure}

To examine phase alignment and amplitude fidelity at high speed, Figure~\ref{fig:temporal-trajectories-4ms} plots the time series of pelvis vertical position (top) and left/right foot horizontal positions (middle and bottom) over 100 time steps at 4 m/s. Solid lines represent the ROM teacher, while dashed lines show the ROM‐GRL policy. The close overlap, along with minimal phase lag and amplitude error, demonstrates that the student policy reproduces both the timing and magnitude of key gait features with high accuracy.

\subsubsection{Quantitative Error Analysis at 4 m/s}

\begin{figure}[htbp]
  \centering
  \includegraphics[width=0.6\linewidth]{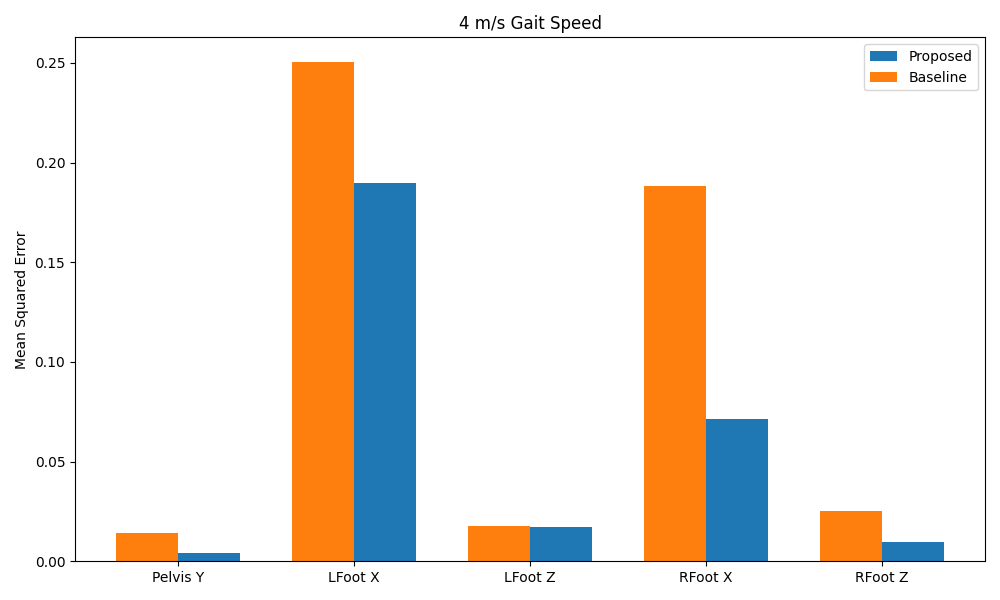}
  \caption{MSE at 4 m/s: pelvis Y and foot X/Z channels, pure‐reward baseline (orange) vs.\ ROM‐GRL (blue).}
  \label{fig:mse-results-4ms}
\end{figure}

We quantify tracking accuracy at 4 m/s by computing the mean squared error (MSE) between each policy and the ROM teacher over five channels: pelvis Y, left/right foot X, and left/right foot Z. Figure~\ref{fig:mse-results-4ms} shows that ROM‐GRL (blue) achieves substantially lower error across all channels compared to the pure‐reward baseline (orange). Table~\ref{tbl:mse-comparison} reports the numerical MSE values and the average percentage improvement: ROM‐GRL reduces pelvis Y error by 71.0\%, left foot X by 24.1\%, left foot Z by 2.4\%, right foot X by 62.1\%, and right foot Z by 62.2\%, for an overall average MSE reduction of 44.4\%.

\begin{table}[htbp]
  \centering
  \caption{MSE comparison at 4 m/s, including percentage reduction for each feature.}
  \label{tbl:mse-comparison}
  \begin{tabular}{lrrr}
    \toprule
    Feature     & Baseline MSE & ROM‐GRL MSE & Reduction (\%) \\
    \midrule
    Pelvis Y    & 0.01407      & 0.00408     & 71.0 \\
    LFoot X     & 0.25035      & 0.18992     & 24.1 \\
    LFoot Z     & 0.01750      & 0.01708     & 2.4  \\
    RFoot X     & 0.18821      & 0.07136     & 62.1 \\
    RFoot Z     & 0.02543      & 0.00962     & 62.2 \\
    \midrule
    \textbf{Average reduction} & \multicolumn{3}{c}{44.4\,\%} \\
    \bottomrule
  \end{tabular}
\end{table}

\subsubsection{Discussion}
Our experimental results at 4\,m/s confirm that incorporating ROM guidance via a simple five‐dimensional discriminator substantially improves humanoid gait tracking. Qualitatively, the ROM‐GRL policy more closely replicates ROM teacher leg trajectories, exhibiting clearer swing‐phase motion and enhanced left–right symmetry. Temporally, phase lag and amplitude mismatches in both pelvis vertical motion and foot horizontal excursions are markedly reduced compared to a pure‐reward baseline. Quantitatively, ROM‐GRL achieves an average MSE reduction of 44.4\,\% across pelvis Y and foot X/Z channels, with individual channel improvements ranging from 2.4\,\% to 71.0\,\%.

To assess generality, we also trained a full‐body policy targeting 1\,m/s using the same 4\,m/s ROM reference data. Despite the differing dynamics at lower speed, the discriminator successfully guided the 1\,m/s policy to produce ROM‐like gait patterns, demonstrating that low‐dimensional ROM outputs capture transferable gait characteristics across speeds without retraining.

Overall, these findings show that even a very compact, dynamically consistent ROM can serve as an effective teacher: by distilling its five key trajectory signals into a full‐body policy via adversarial imitation, we obtain robust, naturalistic walking gaits without any motion capture data or elaborate reward engineering.

\section{Conclusion and Future Work}

In this work, we introduced ROM-GRL, a two-stage reinforcement learning framework that leverages a lightweight, planar 4-DOF ROM as a teacher for full-body humanoid walking. First, we generate energetically efficient reference trajectories from the ROM; Second, we distill those trajectories into a high-dimensional policy via Soft Actor–Critic, augmented by an adversarial discriminator that enforces ROM-style behavior. We achieve natural, stable gaits without any motion capture data. Our experiments at both 1 and 4 meters per second demonstrate that ROM-GRL significantly reduces phase lag and MSE compared to a pure-reward baseline, confirming that even a low-dimensional, dynamically consistent ROM can serve as an effective training guidance.

Despite these successes, ROM‐GRL has several limitations. The current ROM assumes fixed‐length legs and strictly planar dynamics, confining its applicability to a narrow walking regime. Our discriminator operates on just 5-D gait features, which, while effective, may omit subtler aspects of high-dimensional human motion. Moreover, the full‐body policy has been validated only in simulation on flat terrain and has not yet been deployed on hardware or tested in more challenging environments such as slopes, variable friction surfaces, or obstacle-rich settings.

To address these limitations and broaden the scope of our approach, we propose two primary directions for future work. First, we aim to develop variable leg ROMs that incorporate telescoping leg segments such as SLIP-style springs with variable rest lengths, along with actuated leg extension. By modeling leg-length dynamics, these ROMs can generate reference trajectories for running, hopping, and other high‐impact maneuvers, enabling ROM‐GRL to teach a wider range of gaits. Second, we will introduce latent control in the student policy: by conditioning the high-dimensional policy on a low-dimensional latent variable, we can parameterize both target speed and gait style. This enables continuous interpolation between walking, jogging, and sprinting templates without retraining, and provides richer, style-controllable guidance to the agent.

By extending the ROM with variable leg segments and incorporating a latent parameter, we can generate versatile, dynamically consistent gait templates across a wide range of locomotion modes. When paired with curriculum learning over varied terrains and speeds, these templates will guide the training of dynamic, speed-adaptive policies and lay the groundwork for robust sim-to-real transfer.

\bibliographystyle{unsrtnat}
\bibliography{references}  






\end{document}